\documentclass[default,iicol]{sn-jnl}% Math and Physical Sciences Reference Style
\pdfoutput=1 

\usepackage{graphicx}%
\usepackage{multirow}%
\usepackage{amsmath,amssymb,amsfonts}%
\usepackage{amsthm}%
\usepackage{mathrsfs}%
\usepackage[title]{appendix}%
\usepackage{xcolor}%
\usepackage{textcomp}%
\usepackage{manyfoot}%
\usepackage{booktabs}%
\usepackage{algorithm}%
\usepackage{algorithmicx}%
\usepackage{algpseudocode}%
\usepackage{listings}%
\usepackage{subfigure}

\newcommand\scalemath[2]{\scalebox{#1}{\mbox{\ensuremath{\displaystyle #2}}}}

\usepackage{geometry}
\usepackage{colortbl}
\definecolor{best}{rgb}{1,0.6,0.6}
\definecolor{secondbest}{rgb}{1,0.8,0.6}
\definecolor{thirdbest}{rgb}{1,1,0.6}

\newcommand{\rev}[1]{{\textcolor{black}{{#1}}}}

\usepackage[numbers]{natbib}
\newcommand{\algname}{FMGS}
\newcommand{\algfull}{Foundation Model Embedded Gaussian Splatting}

\newcommand{\pos}{\mathbf{x}}

\newcommand{\entriesPerLevel}{E}

\newcommand{\featuresPerEntry}{D}

\newcommand{\levels}{L}
\newcommand{\level}{l}
\newcommand{\resolution}{N}
\newcommand{\minResolution}{N_\mathrm{min}}
\newcommand{\maxResolution}{N_\mathrm{max}}
\newcommand{\perLevelScale}{b}

\newcommand{\nn}{MLP}
\newcommand{\nnClip}{MLP^{CLIP}_{\phi}}
\newcommand{\nnDino}{MLP^{DINO}_{\psi}}

\newcommand{\encOut}{\mathbf{q}}
\newcommand{\clipOutput}{\mathbf{f}}
\newcommand{\featureMap}{\mathbf{F}}
\newcommand{\featureMapDino}{\mathbf{D}}
\newcommand{\dinoOutput}{\mathbf{d}}

\newcommand{\real}{\mathbb{R}}
\newcommand{\cov}{\mathbf{\Sigma}}
\newcommand{\mhe}{MHE}

\newcommand{\improvement}{10.2}

\begin{document}

\title[Article Title]{\algname{}: Foundation Model Embedded 3D Gaussian Splatting for Holistic 3D Scene Understanding}


\author[1]{\fnm{Xingxing} \sur{Zuo}}\email{xingxingzuo@google.com}

\author[1]{\fnm{Pouya} \sur{Samangouei}}\email{samangouei@google.com}

\author[1]{\fnm{Yunwen} \sur{Zhou}}\email{verse@google.com}

\author[1]{\fnm{Yan} \sur{Di}}\email{yanditum@google.com}

\author[1]{\fnm{Mingyang} \sur{Li}}\email{mingyangli@google.com}

\affil[1]{\orgname{Google}}

% \affil[2]{\orgdiv{Department}, \orgname{Organization}, \orgaddress{\street{Street}, \city{City}, \postcode{10587}, \state{State}, \country{Country}}}

% \affil[3]{\orgdiv{Department}, \orgname{Organization}, \orgaddress{\street{Street}, \city{City}, \postcode{610101}, \state{State}, \country{Country}}}


\abstract{
Precisely perceiving the geometric and semantic properties of real-world 3D objects is crucial for the continued evolution of augmented reality and robotic applications. To this end, we present \algfull{} (\algname{}), which incorporates vision-language embeddings of foundation models into 3D Gaussian Splatting (GS). 
The key contribution of this work is an efficient method to reconstruct and represent 3D vision-language models. This is achieved by distilling feature maps generated from image-based foundation models into those rendered from our 3D model. To ensure high-quality rendering and fast training, we introduce a novel scene representation by integrating strengths from both GS and multi-resolution hash encodings (MHE). Our effective training procedure also introduces a pixel alignment loss that makes the rendered feature distance of same semantic entities close, following the pixel-level semantic boundaries.
Our results demonstrate remarkable multi-view semantic consistency, facilitating diverse downstream tasks,
beating state-of-the-art methods by $\mathbf{\improvement}$ percent on open-vocabulary language-based object \rev{detection}, despite that we are $\mathbf{851\times}$ faster for inference. 
This research explores the intersection of vision, language, and 3D scene representation, paving the way for enhanced scene understanding in uncontrolled real-world environments. We plan to release the code on the \href{https://xingxingzuo.github.io/fmgs}{[project page]}.
%  %such as: open-vocabulary language-based object queries and unsupervised open-vocabulary semantic segmentation, 
%The key contribution of this work is a self-supervised distillation process aligns the rendered language feature map, obtained through Gaussian splatting, with 2D feature maps derived from foundation models. %
% \algname{} pioneers a novel fusion of vision, language, and 3D representation, unlocking unprecedented capabilities for object recognition and scene understanding in uncontrolled real-world scenarios.
}




\maketitle

\section{Introduction}\label{sec:intro}

3D scene understanding is a critical task in various computer vision and robotics applications. Yet, most existing methods primarily concentrate on either 3D geometry and appearance estimation~\cite{schonberger2016structure,mildenhall2020nerf,kerbl20233d} or 3D object detection and scene segmentation trained on datasets with closed sets of classes~\cite{scannet,grinvald2019volumetric,narita2019panopticfusion}. However, for an intelligent agent to interact smoothly with the physical world, merely understanding a subset of the space characterized by pre-identified labels is insufficient. Inspired by the latest advancements in foundation models (FMs) with impressive language and vision semantics~\cite{radford2021learning,alayrac2022flamingo}, this paper endeavors to develop a more natural 3D scene representation supporting \rev{open-world visual recognition and understanding}. It integrates both geometric and open-vocabulary semantic information, facilitating easy querying for downstream tasks \rev{such as object detection and semantic segmentation in open-world scenarios}.

In this paper, we utilize Gaussian Splatting~\cite{kerbl20233d} as \rev{the backbone} for reconstructing 3D geometry and appearance, which has demonstrated superior performance in terms of rendering quality \rev{for novel-view image synthesis} and training efficiency. To assist open-vocabulary 3D scene understanding, we rely on pre-train 2D vision-language CLIP~\cite{radford2021learning} and lift the corresponding information into 3D by a novel multi-view training procedure. We note that, in research communities, the system that is most similar to us is LEFR~\cite{kerr2023lerf}, which integrates implicit NERF~\cite{mildenhall2020nerf} based scene representation and CLIP embeddings. Compared to LERF, our system develops a different architecture, provides a variety of technical contributions ranging from high efficiency to 3D consistent query, and obtains significantly better results (approximately $\improvement$ percent in representative key metrics).

A straightforward approach to enhance 3D Gaussian Splatting with vision-language FM embeddings is to attach each Gaussian with a learnable feature vector, \rev{which can be trained} through image rasterization to formulate loss functions. However, maintaining high-quality rendering with GS typically requires millions of Gaussians in a nominal room-scale environment. Employing per-Gaussian feature vectors inevitably results in excessive memory consumption and significantly slows down training, limiting the practical applications of this system. Motivated by iNGP~\cite{muller2022instant}, we model our system by using 3D Gaussians together with multi-resolution hash encoding (MHE) to distill the foundation model embeddings. Specifically, to obtain the language embedding from the Gaussians, we utilize their mean values to query the MHE field at corresponding positions. Subsequently, this queried MHE is processed through a Multi-Layer Perceptron (MLP) to generate the output language embedding.

In the training phase, we employ a supervision mechanism on the MHE-based language FM CLIP feature field using a hybrid feature map. This map is derived from the average of multi-scale image crops obtained from various viewpoints. This approach enables the embedding to effectively capture language features corresponding to each scale ensuring a comprehensive representation. For instance, the embedding might represent a `red book' when viewed up-close, while depicting a `library' from a more distant perspective. It is noteworthy that CLIP embeddings are designed to encapsulate the overall concept presented in a 2D image, exhibiting minimal variation across individual pixels. Additionally, CLIP embeddings are not perfectly multi-view consistent, i.e., when a 3D object observed by a moving camera via different views, the difference between computed CLIP embeddings across frames are not explicitly minimized.
To solve the above-mentioned problems, we rely on multi-view consistency training process to ensure that 3D models, when rendered from different image views, exhibit minimal variations. Additionally, to allow pixel-aligned query experience, DINO~\cite{dino} embeddings are used together with CLIP embeddings similar to LERF~\cite{kerr2023lerf}. By carefully analyzing the properties in both CLIP and DINO embeddings, we design an additional pixel alignment loss to further improve the object localization and scene understanding capabilities. This loss is grounded in the dot product similarity of CLIP/DINO features between the central pixel and its surroundings, guiding the rendered CLIP feature map to replicate the same similarity pattern observed in the DINO feature map.

%\begin{figure}[h]
%\centering
%\includegraphics[width=0.5\textwidth]{}
%\caption{Overview figure placeholder.\pouya{Rocky please add this}}
%\end{figure}

%Once \algname{} is trained we can create 3D relevancy maps for common and long-tail concepts. We evaluate \algname{} on the scenes that are captured using regular smartphones and show that we can localize both specific queries such as a "wallet" and more conceptual queries such as "furniture". We also perform quantitative comparisons against state-of-the-art method \cite{kerr2023lerf} and show we improve by more than $10\%$ on localization accuracy.

This research paves the way for enhanced real-world applications, such as augmented reality experiences where users can interact with objects using natural language and robotic systems that can navigate and manipulate environments based on linguistic commands. By bridging the gap between language and 3D representation, \algname{} opens up new possibilities for understanding and interacting with our surroundings.

%\rocky{The introduction section needs to be significantly further extended, we should describe of the motivation of our contributions as well.}\pouya{done, please review} 

Our contributions can be summarized as follows:
\begin{itemize}
\item Novel semantic scene representation: We introduce a novel approach combining 3D Gaussians (parameterized by mean, covariance, opacity, and spherical harmonics) for geometry and appearance representation, with MHE for efficient \rev{semantic} embedding. This approach addresses memory constraints in room-scale scenes including millions of 3D Gaussians.
\item Multi-view consistent language embeddings: 
Our training process utilizes Gaussian-splatting based rendering from multiple views, ensuring consistency across \rev{3D space in static scenarios}. Language embeddings remain invariant to viewpoints, enforcing local proximity consistency within Gaussian volumes.
\item Addressing pixel misalignment: We address pixel alignment challenges of CLIP features by extracting and aggregating them at multiple resolutions for a hybrid CLIP feature, \rev{which is used for} supervising the training. Regularization with pixel-aligned DINO features and a novel dot-product similarity loss enhances spatial precision and object differentiation. 
\item State-of-the-art performance: 
Our methods demonstrate superior performance in open-vocabulary semantic object localization, outperforming existing state-of-the-art approaches with quantitative and qualitative results by a wide margin, despite being hundreds of times faster.
\end{itemize}

\section{Related Works}
We review three main areas of related articles: 3D scene representation, open-vocabulary object recognition and scene understanding, and combined 3D scene representation and semantic understanding.

\paragraph{3D Scene Representation}
Scene representation in 3D can be roughly categorized by mesh based, voxel based, point based, and implicit ones.
Voxel based methods typically discretize 3D space into regular grid cell elements where each grid cell corresponds to a voxel. To estimate the dense 3d voxel cells, probabilistic fusion methods were firstly~\cite{izadi2011kinectfusion} used and researchers also developed end-to-end learn-able methods~\cite{sun2021neuralrecon}, by using either depth sensors~\cite{izadi2011kinectfusion} or monocular camera systems~\cite{yang2020mobile3drecon}. 
To visualize estimated voxel fields, they are typically converted into a mesh based representation. 
%
% This allows efficient rendering on modern computer graphic systems. 3D meshes can also be constructed by alternative methods~\cite{schops2019surfelmeshing,lin2021end}. Although both voxel and mesh based methods have achieved significant success in a variety of fields, their nature of discrete scene representation limits the capabilities of novel view synthesis and photo-realistic reconstruction and rendering performance. 
%
This enables efficient rendering on modern computer graphics systems. While alternative methods, such as those using 3D meshes~\cite{schops2019surfelmeshing,lin2021end}, have achieved notable success in various fields, their discrete scene representation, whether voxel-based or mesh-based, imposes limitations on the ability to achieve photo-realistic reconstruction and rendering performance.

Neural implicit representation, e.g., NeRF series~\cite{mildenhall2020nerf,mipnerf,mipnerf360, barron2023zipnerf}, represent 3D scenes by fully-connected neural networks, in which volume density and radiance can be queried by input position and view direction vectors. To improve the training and rendering efficiency of NeRFs, 3D space can be discretized by using MHE similar to the concept used in voxel based methods~\cite{muller2022instant}.
TensoRF \cite{chen2022tensorf} models radiance fields as 4D tensors, factorizing them into compact low-rank tensor components using CP decomposition and introducing novel vector-matrix (VM) decomposition for improved rendering quality, reduced memory footprint, and faster reconstruction. 

Finally, point-based methods are originally widely used for directly processing data from depth sensors, for performing geometrical and semantic computer vision tasks~\cite{qi2017pointnet,karkus2021differentiable}.  Point-NeRF~\cite{point-nerf} efficiently combines point cloud and NeRF to achieve impressive fast view synthesis results. Recently, 3D Gaussian Splatting (GS) has been proposed to model points as 3D Gaussians for scene representation~\cite{kerbl20233d}, and achieved state-of-the-art novel view synthesis rendering quality. However, in~\cite{kerbl20233d}, the number of Gaussians used for scene representation can easily surpass one million, which introduces strict memory and computational requirements for downstream use cases. 

%DVGO \cite{sun2022direct} and Plenoxels \cite{fridovich2022plenoxels} optimize voxel grids of (neural or SH) features for fast radiance field reconstruction. EG3D \cite{chan2022efficient} uses a tri-plane representation for 3D GANs. Instant-NGP \cite{muller2022instant} uses multi-resolution hashing for efficient encoding and also leads to high compactness. \algname{} borrows the Instant-NGP architecture on top 3D Gaussians to efficiently encode the language embeddings.

\paragraph{Open-Vocabulary Object Detection and Scene Understanding}
Advancements in open-vocabulary object detection in 2D images have been made by leveraging natural language prompts. LSeg~\cite{li2022language_lseg} employs a text encoder for semantic label embeddings and a transformer-based image encoder for dense pixel embeddings, using contrastive alignment to achieve zero-shot image segmentation and generalization to unseen categories. CRIS~\cite{wang2022cris} leverages CLIP for image segmentation, employing a vision-language decoder to align text and pixel-level features, and text-to-pixel contrastive learning to enforce similarity between text and relevant pixel features. CLIP-Seg~\cite{luddecke2022image_clipseg} leverages CLIP as a backbone, employs a transformer-based decoder for dense prediction, and generates image segmentation based on arbitrary text or image prompts. OV-Seg~\cite{liang2022open_ovseg} improves open-vocabulary semantic segmentation by finetuning CLIP on masked image regions and text descriptions from noisy captions, achieving promising performance without dataset adaptations.

Current approaches often employ region proposal or mask prediction methods to guide open-vocabulary classification models. OpenSeg~\cite{ghiasi2021open_openseg} employs mask representations to facilitate visual grouping and align captions with predicted segmentation masks for open-vocabulary image segmentation. ViLD~\cite{gu2021open_vild} advances open-vocabulary object detection by distilling knowledge from a pretrained image classification model (teacher) into a two-stage detector (student), aligning region embeddings of detected boxes with text and image embeddings inferred by the teacher. Detic~\cite{zhou2022detecting_detic} expands object detectors' vocabulary by training their classifiers on image classification data, outperforming prior methods on open-vocabulary and long-tail detection benchmarks, achieving generalization to new datasets without finetuning and enabling detectors trained on all ImageNet classes. OVIR-3D \cite{lu2023ovird} enables open-vocabulary 3D object instance retrieval by fusing text-aligned 2D region proposals into 3D space, leveraging 2D datasets.

% \algname{} takes inspiration from these methods but operates in a 3D space. It avoids region proposals by incorporating language embeddings into a dense, multi-scale field, enabling hierarchical text queries. Additionally, \algname{} addresses the integration of language into 3D scene understanding, similar to recent works in point cloud processing. 
Open-vocabulary scene understanding has also been explored by using point cloud as sensor inputs. 
PointCLIP~\cite{zhang2021pointclip} aligns CLIP-encoded point cloud with 3D category texts, transferring knowledge from 2D to 3D recognition by projecting point cloud into multi-view depth maps, using an inter-view adapter for global feature extraction and few-shot knowledge fusion. ULIP series~\cite{xue2022ulip,xue2023ulip2} learn a unified representation for images, texts, and 3D point cloud by leveraging pre-trained vision-language models and automatically synthesized triplets, improving the performance of various 3D backbones.
%ULIP-2 \cite{} improves 3D representation learning through tri-modal pre-training, leveraging large multimodal models to generate language descriptions for 3D objects without requiring 3D annotations, enhancing scalability and comprehensiveness.
Lu et al.~\cite{lu2023open} leverage pre-trained image and vision-language models and cross-modal contrastive learning for open-vocabulary 3D point cloud detection without 3D annotations. 
%~\cite{ding2022language}, but focuses on optimizing scene representation from multi-view images.

\paragraph{Combined 3D Scene Representation and Semantic Understanding}

Language has been incorporated into 3D scene understanding in various ways.
For the task of visual question answering, systems like iQA \cite{gordon2018iqa}, ScanQA \cite{azuma2022scanqa}, and SimVQA \cite{cascante2022simvqa} leverage 3D information to answer queries about the environment. For object recognition enhancement, language and shape information can be combined to improve object recognition, as seen in \cite{corona2022voxel} and \cite{thomason2022language}. 

%Incorporating language guidance into neural rendering poses challenges.
Inspired by the success of implicit neural reconstruction~\cite{mildenhall2020nerf,mipnerf,mipnerf360}, researchers also start to explore incorporating language guidance into 3d neural scene representation. 
LERF~\cite{kerr2023lerf} enables open-ended language queries in 3D by incorporating language embeddings from models, e.g. CLIP, into NeRF. 3D-OVS~\cite{liu20233d} leverages pre-trained CLIP and DINO models in a weakly supervised manner, distilling multi-modal knowledge and object reasoning into a neural radiance field (NeRF) for segmentation task. 

Tschernezki et al.~\cite{tschernezki22neural} leverage a pre-trained 2D image feature extractor to train a 3D student network, boosting performance in analyzing multiple images forming a 3D scene. 
FFD~\cite{kobayashi2022distilledfeaturefields} tackles scene editing by distilling knowledge from pre-trained 2D image feature extractors into a 3D feature field that guides local editing based on user queries. 
VL-Fields~\cite{tsagkas2023vlfields}, a neural implicit spatial representation fusing scene geometry and vision-language features, enables open-vocabulary semantic queries without requiring prior object class knowledge. FeatureNeRF~\cite{ye2023featurenerf} distills pre-trained vision models (DINO, Latent Diffusion) to learn generalizable NeRFs, leveraging neural rendering for 2D-to-3D mapping and extracting deep features from NeRF MLPs. 

Additionally, ConceptFusion~\cite{conceptfusion} enables open-set and multimodal reasoning in 3D scene representations by fusing foundation model features with SLAM and multi-view fusion.
ConceptGraphs~\cite{conceptgraphs} leverages 2D foundation models and multi-view association to capture semantic and spatial relationships for efficient task-driven planning. OpenMask3D~\cite{takmaz2023openmask3d} aggregates per-mask features using the multi-view fusion of CLIP-based image embeddings guided by predicted class-agnostic 3D instance masks. 
SA3D~\cite{cen2023segment} enables 3D segmentation of target objects in neural radiance fields (NeRF) through one-shot manual prompting, leveraging density-guided inverse rendering, cross-view self-prompting, and an iterative process to project 2D segmentation masks onto 3D mask grids.
PVLFF~\cite{chen2023panoptic} generates a scene's feature field, combining vision-language and hierarchical instance features through contrastive loss from 2D instance segment proposals.

CLIP-Fields~\cite{shafiullah2022clip} learns a spatial mapping to semantic embeddings via weak supervision from web-trained language and vision models, enabling tasks like object identification and robot navigation without direct human labeling.
GNFactor~\cite{Ze2023GNFactor}, a multi-task robotic manipulation agent, leverages a shared 3D voxel representation and language-augmented neural fields for generalizable visual behavior cloning. 

Our work is close and directly comparable to LERF~\cite{kerr2023lerf} in terms of assumptions about information available at training phase and query time. For example, it does not assume a priori knowledge of query categories at training time which is assumed 3D-OVS~\cite{liu2023_3dovs}.

\rev{
Recently, several concurrent works have emerged, investigating the distillation of semantic features into GS scene representations~\cite{chen2024survey,fei20243d}.
%
%After the initial submission of this paper, several research articles have emerged addressing similar problems, and survey papers~\cite{chen2024survey,fei20243d} also offer comprehensive overviews of related fields.
%
LEGaussians~\cite{shi2023language} introduces a method for quantizing high-dimensional concatenated CLIP and DINO features into compact ones to conserve memory, and attach to each individual Gaussian.
Langsplat~\cite{qin2023langsplat} segments images by SAM~\cite{kirillov2023segment} and then inputs hierarchical semantic segments into CLIP to extract semantic features for the segments. To address memory constraints, Langsplat incorporates a scene-specific language autoencoder to encode CLIP features into lower dimensions, which are also attached to each individual Gaussian.
Feature 3DGS~\cite{zhou2023feature}  distills pixel-aligned features
from 2D foundation models, such as SAM~\cite{kirillov2023segment}
and LSeg~\cite{li2022language_lseg}, into GS by associating each Gaussian with a learnable vector.
%which is then upsampled using a $1 \times 1$ convolution layer. 
However, this suffers from a loss of semantic understanding capability, particularly for long-tail semantics~\cite{kerr2023lerf}.
Different from all the aforementioned methods, which attach semantic features to each Gaussian, we propose to seamlessly integrate 3D Gaussian scene representation together with MHE for efficient semantic encodings. Notably, our method does not necessitate additional scene-specific quantization or auto-decoder steps, thereby preserving the semantic features' representation capability from foundation models. 
%Additionally, we propose dedicated methods to address the pixel-misalignment issue of CLIP.
}

\section{Background Methods}

\subsection{3D Gaussian Splatting}
GS \cite{kerbl20233d} represents an environment using a set of 3D Gaussians, each defined by a mean $\mathbf{\mu} \in \real^3$, an anisotropic covariance matrix $\cov \in \real^{3\times3}$, an alpha value $\alpha \in [0, 1]$ representing opacity, and spherical harmonics coefficients (SH). Given a 3D position $\mathbf{\pos} \in \real^3$, the probability density function of 3D Gaussian is defined as:

\begin{equation}
\label{eq:gaussian}
G(\mathbf{\pos})~= e^{-\frac{1}{2}(\mathbf{\pos} - \mathbf{\mu})^{T}\cov^{-1}(\mathbf{\pos} - \mathbf{\mu})}
\end{equation}
where $(\cdot)^{T}$ represents a transpose operation and $(\cdot)^{-1}$ denotes matrix inversion.
To render 3D Gaussians in 2D, we project their mean positions by point projection, and project their covariance using the following equation:

\begin{equation}
\label{eq:projection}
\cov' = \mathbf{J} \mathbf{W} ~\cov ~\mathbf{W} ^{T}\mathbf{J}^{T}
\end{equation}
where $\mathbf{W} \in \real^{3\times3}$ is the viewing transformation and $\mathbf{J}\in \real^{3\times3}$ is the Jacobian of the affine approximation of the projective transformation~\cite{zwicker2001ewa}. To optimize covariance matrices, we use an equivalent representation:

\begin{equation}
\label{eq:optimization}
\cov = \mathbf{R}\mathbf{S}\mathbf{S}^T\mathbf{R}^T
\end{equation}
where $\mathbf{R} \in \real^{3\times3}$ and $\mathbf{S} \in \real^{3\times3}$ are rotation and scaling matrices, respectively. GS also includes spherical harmonics coefficients to model the appearance of the scene. Gradients for all parameters are derived explicitly to avoid overhead during training.

Each Gaussian encodes the color $c$ using spherical harmonics, which gives a value depending on the viewing directions. The $\alpha-$blending point-based rendering for a pixel color $\mathbf{c}$ is done by blending $\mathcal{N}$ points in the depth order from front to back:
\begin{equation}
\mathbf{c} = \sum_{i \in \mathcal{N}}
\mathbf{c}_{i}\alpha_{i}
\prod_{j=1}^{i-1}(1-\alpha_{j}),
\label{eq:render}
\end{equation}
where $\alpha_i$ is given by a 2D Gaussian multiplied by a learned per Gaussian opacity \cite{yifan2019differentiable}.

Note that although the image rendering model is similar across NeRFs and GS, the rendering algorithm is much more efficient in GS. NeRFs need to march along the ray to integrate volume, however, \rev{GS rendering} uses a point-based $\alpha-$blending approach. This allows GS to include a real-time rendering solution that leverages GPU sorting algorithms and draws inspiration from tile-based rasterization. By using a 3D Gaussian representation, anisotropic splatting can be performed while respecting visibility order. This is achieved through sorting and alpha-blending. Additionally, a fast and accurate backward pass is enabled by tracking the traversal of sorted splats.

\begin{figure*}[tbh]
    \centering
    \includegraphics[width=\textwidth]{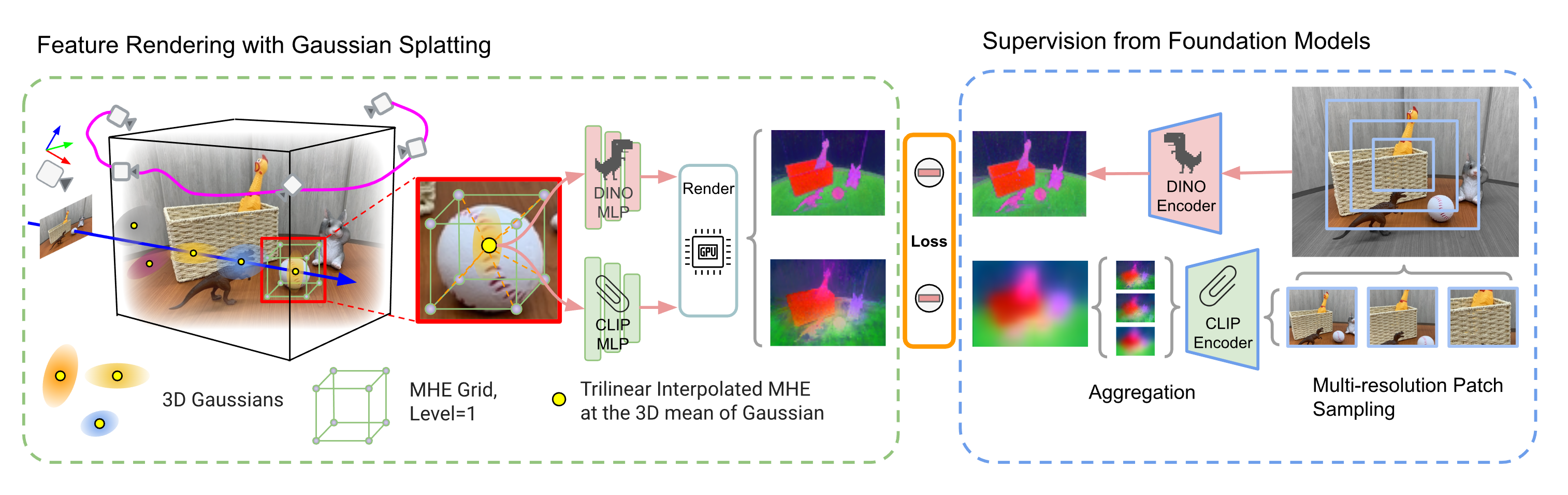}
    \caption{
    \textbf{\algname{} Training pipeline:} 
    \textit{Left}: Shows how \algname{}' feature field renders \rev{CLIP and DINO feature maps} for loss calculation. The feature field is a multi-resolution hash encoder (MHE)~\cite{muller2022instant} that \rev{embeds semantic information} into 3D Gaussians acquired from 3D Gaussian Splatting \cite{kerbl20233d}. \textit{Right}: Shows the target DINO feature map and hybrid CLIP feature map from the foundation models. Note, \rev{for visualization simplicity}, we \rev{only show} a single-level MHE here but in implementation we have used multiple levels and concatenate their encodings.}  
    \vspace{-5pt}
    \label{fig:training}
\end{figure*}

\subsection{Multi-resolution Hash Encoding}
Representing a 3D feature field can have many forms. 
A naive method is to attach a feature vector (or multiple) to each Gaussian, which can be optimized along with other Gaussian parameters (position, covariance, and so on). However, this is extremely costly in terms of computational cost and memory consumption especially when a large number of Gaussians are generated for scene representation. 
%In fact, adding a $512\times1$ feature vector per Gaussian will increase the number of optimized parameters to be $9.98\times$ under authentic GS parameterization~\cite{kerbl20233d} (9 geometric parameters and 48 spherical harmonic appearance parameters) and $65.0\times$ under simplified GS parameterization~\cite{keetha2023splatam} (5 geometric parameters and 3 color appearance parameters).
In fact, adding a $512\times1$ feature vector per Gaussian will increase the number of optimized parameters to be \rev{$9.83\times$ under authentic GS parameterization~\cite{kerbl20233d} (10 geometric parameters and 48 spherical harmonic appearance parameters per Gaussian) and $65.0\times$ under simplified GS parameterization~\cite{keetha2023splatam} (4 geometric parameters and 4 appearance parameters per Gaussian)}.

To mitigate this problem, we are motivated by multi-resolution hash embedding (MHE)~\cite{muller2022instant}, which provides efficient scene representation that consists of two trainable components. The first component first hashes a given position $\mathbf{\pos} \in \real^3$, and then looks up into a trainable hash table for the corresponding embedding. The second component is an MLP that takes the corresponding embeddings and makes predictions such as color and density. The representation contains multiple hash tables, one per each scale.
Specifically, MHE first encodes a given position $\encOut=\mhe_\theta(\mathbf{\pos})$. To do so, it contains a hash table with $\levels$ levels. Each level contains up to $\entriesPerLevel$ feature vectors with dimensionality $\featuresPerEntry$. Resolution of each level is determined by $\resolution_\level = \left\lfloor \minResolution \cdot \perLevelScale^\level \right\rfloor$
where $\minResolution$ is the coarsest resolution, $\maxResolution$ is the finest resolution, and $\perLevelScale$ is a growth factor.

To get $\encOut$ for a given position $\mathbf{\pos}$, we query MHE at all scales and concatenate the resulting features. For each scale, we find the enclosing voxel for $\mathbf{\pos}$. Then, each corner entry of the voxel is mapped into a feature vector \rev{with dimensionality $\featuresPerEntry$  according to the trainable hash table}. 
%
%Note that if the scale's grid has a larger number of such corners than $\entriesPerLevel$ collision happens. Different resolution levels complement each other. Coarser levels are injective (no collisions) but offer only low-resolution representation. Finer levels can capture small features but suffer from collisions. Gradients of more important samples dominate the collision average, leading to an effective resolution of hash collisions. The downstream component learns to resolve these collisions as well.  \rocky{Double check this!}
%
MHE trilinearly interpolates the queried corner entries according to their relative position of $\mathbf{\pos}$  in its hypercube for each level. This ensures the continuity of the encoded input and its composition with the neural network, avoiding grid-aligned discontinuities and blocky appearance.
After this mapping is done, the features from all scales are concatenated to each other, and the auxiliary inputs $\psi \in \real^K$ which results in a feature vector $\encOut$ of size $\levels \times \featuresPerEntry + K$. The resulting encoding then goes to the second component which is an MLP network, $\nn_\Phi(\encOut)$, produces the final output.
This architecture significantly reduces the number of weights that are trained for each view while having an $O(1)$ GPU look up for hashing. Overall this results in significant improvements in quality and speed of training.

\section{Method}
Our method, i.e. \algfull{} (\algname{}), leverages strengths of both GS and MHE. We rely on GS for efficient and accurate scene geometry representation and on MHE for representing the scene's language content in a light-weighted manner. Given a set of input images, we compute \rev{the corresponding camera poses and 3D sparse visual points} using an off-the-shelf structure from motion system, e.g., COLMAP~\cite{schonberger2016structure_colmap}.  After that we train GS and acquire 3D Gaussians.

Subsequently, we train \rev{the feature embedding field (MHE)} in 3D by grounding 2D CLIP embeddings. This requires us to generate pixel-aligned features on a set of calibrated input images. However, CLIP embeddings are global in nature and not suitable for pixel-aligned feature extraction. To overcome this challenge, we introduce a framework to learn a volumetric language embedding field that embeds over the 3D Gaussians. The field effectively generate features that is the average CLIP features across all views that include that 3D Gaussian. To supervise our dense feature field, we create a hybrid feature map based on CLIP embeddings across multi-scale crops of training views. % Our approach provides a significant enhancement over LERF's NeRF-based approach. 
Figure~\ref{fig:training} provides an overview of our training pipeline.

\subsection{Feature Field Architecture}~\label{sec:ffa}
3D Gaussian Splatting produces millions of Gaussians to enable high quality rendering of a room-scale scene. This makes it very inefficient to have one CLIP feature per Gaussian since these features are high dimensional and keeping all of these features in GPU memory is not feasible.

\rev{To this end}, we parameterize our feature field efficiently using MHE. For a given 3D Gaussian $G(\mathbf{\pos})$ with mean position $\pos$, we first encode $\mathbf{\pos}$ to a feature vector $\encOut=\mhe_\theta(\mathbf{\pos})$ where $\theta$ is our multi-resolution hash table parameters. We subsequently feed this output into an MLP, which generates our language embedding $\hat{\clipOutput} = \nnClip(\encOut)$, with $\hat{\clipOutput}$ belonging to $\mathbb{R}^D$. \rev{We also normalize $\hat{\clipOutput}$ to make it a unit vector.}

\begin{figure*}
    \centering
    \includegraphics[width=\textwidth]{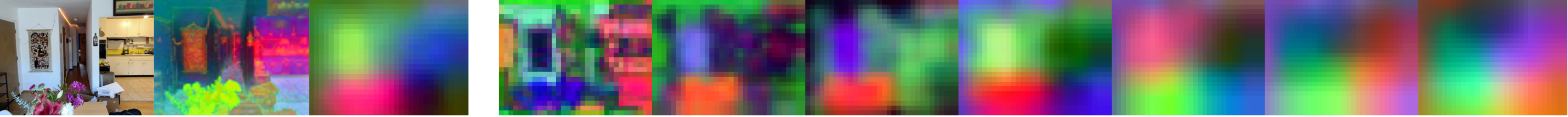}
    \caption{\textbf{The features extracted from foundation models.} The left three subfigures include the RGB image, extracted DINO features from the foundation model, and the hybrid CLIP feature, which is an average of multi-scale CLIP feature maps shown on the right. On the right, the shown seven CLIP feature maps are the extracted from an image pyramid at multiple scales using the foundation model. The resolution of CLIP features decreases from left to right.} 
    \label{fig:blurred_CLIP}
\end{figure*}

\begin{figure*}[tbh]
    \centering
    \includegraphics[width=0.7\textwidth]{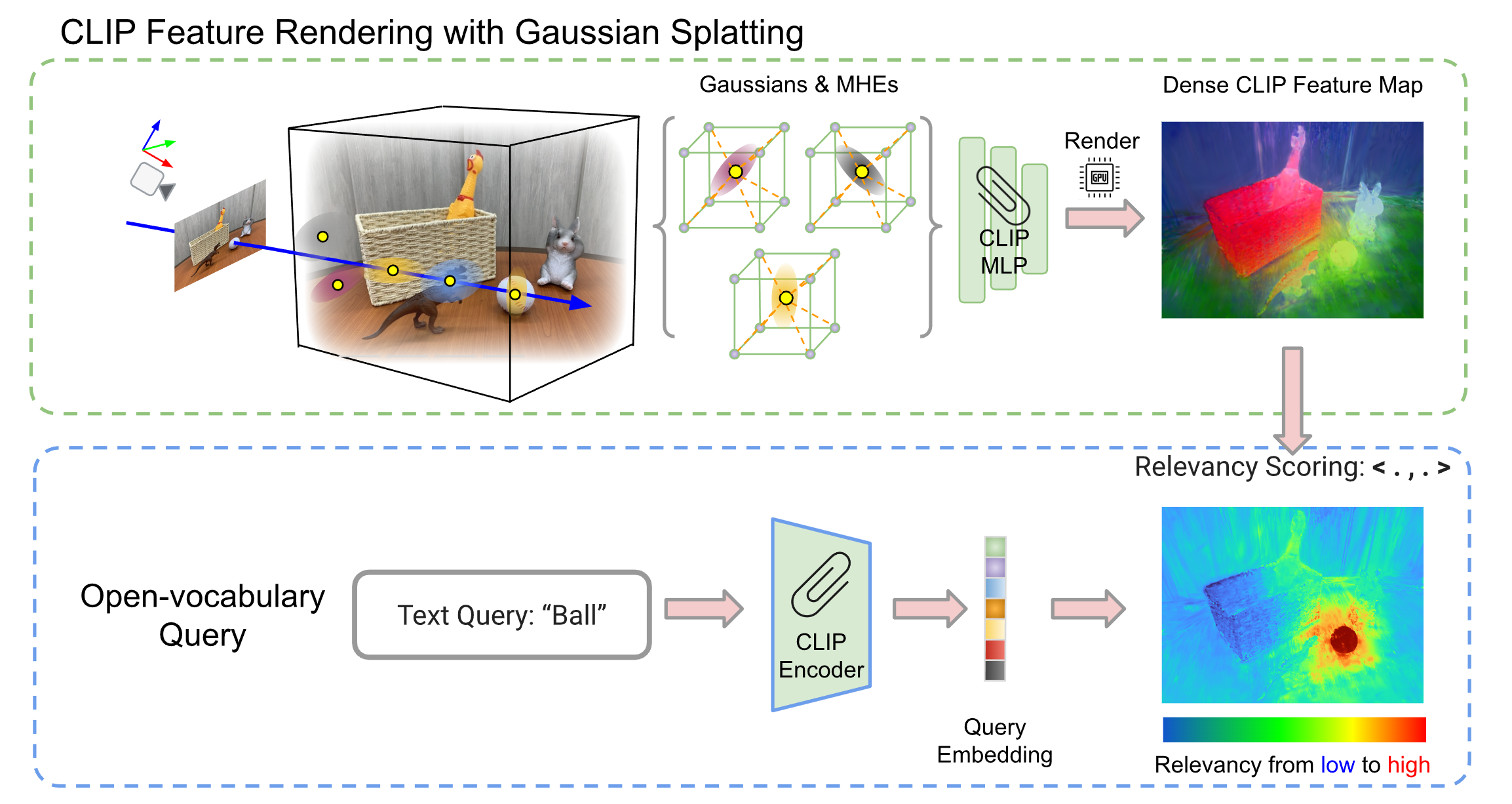}
    \caption{
    \textbf{\algname{} Query pipeline:} \textit{Top}: Given a query view to localize a query, \algname{} first renders the dense CLIP feature map. \textit{Bottom}: given an open-vocabulary query, \algname{} generates a relevancy map highlighting the relevant part of the rendered CLIP feature map to the query embedding. The highest relevant is colored as red while the lowest relevant part is colored as blue. Note, \rev{for visualization simplicity}, we \rev{show a single-level MHE in this figure while used multiple in implementations}.
    }
    \vspace{-1pt}
    \label{fig:query}
\end{figure*}

\subsection{Embed the Foundation Models}
We embed the semantic embeddings from foundation models to our scene representation. Training the semantic embedding has three aspects. First, we use our scene representation to render a predicted feature map $\hat{\featureMap} \in \real^{W \times H \times D}$ where $W$ is the width, $H$ is the height, and $D$ is the dimension of the feature map. Second, we generate a target feature map $\featureMap$ by feeding the view to a FM. Finally we need to ensure that the predicted feature map is aligned with the corresponding target pixels and follows the same object boundaries in terms of feature similarity.

\paragraph{Hybrid CLIP Feature for Supervision}
To supervise our feature field outputs, given a calibrated input image, we first rasterize the features into a 2D feature map $\hat{\featureMap}$ where the $(i, j)$th feature is acquired by point-based $\alpha-$blending:
\begin{equation}
\hat{\clipOutput}_{i, j} = \sum_{k \in \mathcal{N}}
\hat{\clipOutput}_{k}\alpha_{k}
\prod_{l=1}^{i-1}(1-\alpha_{l})
\end{equation}
To generate our target CLIP feature map, denoted as $\featureMap$, we initially pre-compute a multi-scale feature pyramid of CLIP embeddings, similar to the approach used in LERF~\cite{kerr2023lerf}. This involves feeding image patches at various sizes into the CLIP foundation model.
However, in contrast to LERF, which trains its scene representation by interpolating embeddings from the pre-computed CLIP feature pyramid at random scales, we rely on a single hybrid CLIP feature map for training our scene representation.
We scale up the embeddings of the smaller scales in the pre-computed CLIP feature pyramid bilinearly to the largest scale feature map, and generate the hybrid feature map by averaging them. We define our CLIP loss by the following Huber loss:
% \begin{equation}
% \mathcal{L}_{CLIP} = |\hat{\featureMap} - \featureMap|
% \label{eq:CLIP_loss}
% \end{equation}
\rev{
\begin{equation}
\mathcal{L}_{CLIP} = \begin{cases} 
0.5 | \hat{\featureMap} - \featureMap |^2, & \text{if } |\hat{\featureMap} - \featureMap| < \delta \\
\delta \cdot (|\hat{\featureMap} - \featureMap| - 0.5 \cdot \delta), & \text{otherwise}
\end{cases}
\end{equation}}
\rev{where $\delta$ is a hyperparameter, which is set to be $1.25$ empirically.}
As seen in Figure \ref{fig:blurred_CLIP} \rev{where we use PCA to visualize feature maps following FFD~\cite{kobayashi2022distilledfeaturefields}}, we notice that the target CLIP feature map is not fine-grained enough when embedding similarities of neighboring pixels are considered. This results in poor pixel-alignment gradient signals on Gaussians that are not relevant semantically. \rev{On the other hand}, DINO \cite{dino} features give sharp boundaries between objects \cite{amir2021deep} in terms of embedding similarity,~\rev{which can be used for additional regularization}.

\paragraph{Regularization with DINO Feature}
To transfer the characteristic of DINO features while maintaining the CLIP embedding semantics, we (a) add \rev{a} DINO feature field loss and (b) define a pixel-alignment loss between the DINO and CLIP feature fields. The DINO feature field shares the same hash grid parameters as CLIP and gives the same encoding $\mathbf{q}$ for a given $\pos$. Then the DINO feature field outputs $\hat{\dinoOutput} = \nnDino(\mathbf{q})$ where $\psi$ denotes the parameters of the MLP that are not shared with $\nnClip$. This feature field is supervised by passing the $sampled\, image$ once to the pre-trained DINO model  without scaling, yielding $\featureMapDino \in \real ^{W \times H \times L}$ where $L$ is the DINO feature dimension. We then render $\hat{\featureMapDino}$ using the same approach as rendering $\hat{\mathbf{F}}$. The DINO regularization loss is as follows:
\rev{
\begin{equation}
% \mathcal{L}_{DINO} = |\hat{\featureMapDino} - \featureMapDino| 
\mathcal{L}_{DINO} = | \hat{\featureMapDino} - \featureMapDino |^2
\label{eq:DINO_loss}
\end{equation}
}

\paragraph{Pixel-alignment with Dot Product Similarity}
We define a pixel-alignment loss by defining a kernel around every pixel and enforce the dot product similarity \rev{in normalized embedding spaces (between DINO and CLIP) are consistent across} the center pixel and \rev{surrounding ones}. We normalize both rendered features to unit norm, \rev{and then compute the loss}:
\rev{
\begin{equation}
\mathcal{L}_{pixel} = \frac{1}{K^2 -1}  \sum_{\substack{i \in \mathcal{P}}} \sum_{\substack{j \in \mathcal{N}(i), \\ j \neq i}} |\hat{\dinoOutput}_{i}^T\hat{\dinoOutput}_{j} - \hat{\clipOutput}_{i}^T\hat{\clipOutput}_{j}|  
\end{equation}}
\rev{where $\mathcal{P}$ denotes the set of all the pixels in the image, and $\mathcal{N}(i)$ is the $K \times K$ patch kernel around the rendered feature at pixel $i$.}
This makes the rendered CLIP feature follow the same similarity pattern as the DINO feature. Note that we stop the gradient back-propagation through \rev{the rendered} DINO features in this training loss, which means $\nnDino$ would not be affected by this loss. \rev{$\mathcal{L}_{pixel}$ is also termed as ``dotsim" loss for the rest of the paper, since it is formulated by dot product similarity.}

% \begin{equation}
% \mathcal{L}_{pixel} = \frac{1}{K^2 -1}  \sum_{\substack{i \in \mathcal{P} \\ where}} \sum_{\substack{j \in \mathcal{N} \\ i \neq j}} |\hat{\dinoOutput}_{i}^T\hat{\dinoOutput}_{j} - \hat{\clipOutput}_{i}^T\hat{\clipOutput}_{j}|  
% \end{equation}
% where $\mathcal{P}$ is over all 2D positions and $\mathcal{N}$ is the $K \times K$ kernel around each rendered feature excluding that feature. 

\begin{figure*}[tbh]
    \centering
    \includegraphics[width=\textwidth]{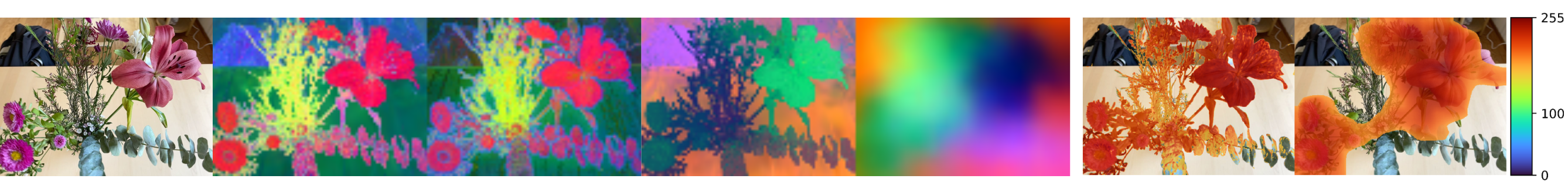}
    \vspace{-12pt}
    \caption{\textbf{Features for Training and Rendered \rev{Views}.} \textbf{Left}: From left to right, the figures show the RGB image, the rendered DINO feature map, the raw DINO feature map extracted for training, the rendered CLIP feature map, and the raw CLIP feature map used for training. \textbf{Right}: We display the relevancy scores for the rendered and raw CLIP feature maps with the text query `flower', where the color bar indicates relevancy scores normalized within the 0-255 range. Notably, querying  the raw CLIP feature map is much inferior to querying the rendered CLIP feature map.}
    \label{fig:raw_CLIP_relevancy}
\end{figure*}

\paragraph{Training Loss}
Overall our total loss is
\rev{
\begin{equation}
    \mathcal{L}_{total} = \lambda \mathcal{L}_{CLIP} + (1-\lambda) \mathcal{L}_{DINO} + \gamma \mathcal{L}_{pixel}
\end{equation}
}
We take \rev{the mean reduction over all the pixels in the image plane} when computing different loss terms. We also empirically find out adding $\mathcal{L}_{pixel}$ in later iterations during training produces the best results. In Figure~\ref{fig:raw_CLIP_relevancy}, we provide examples of features extracted from foundation models for training and the rendered features generated by our trained hybrid semantic scene representation. It is evident that the rendered feature maps exhibit higher quality when compared to the \rev{raw feature maps} obtained directly from the foundation models, owing to our training process enforces multiple-view consistency. 

\begin{figure*}[tbh]
    \centering
    \includegraphics[width=\textwidth]{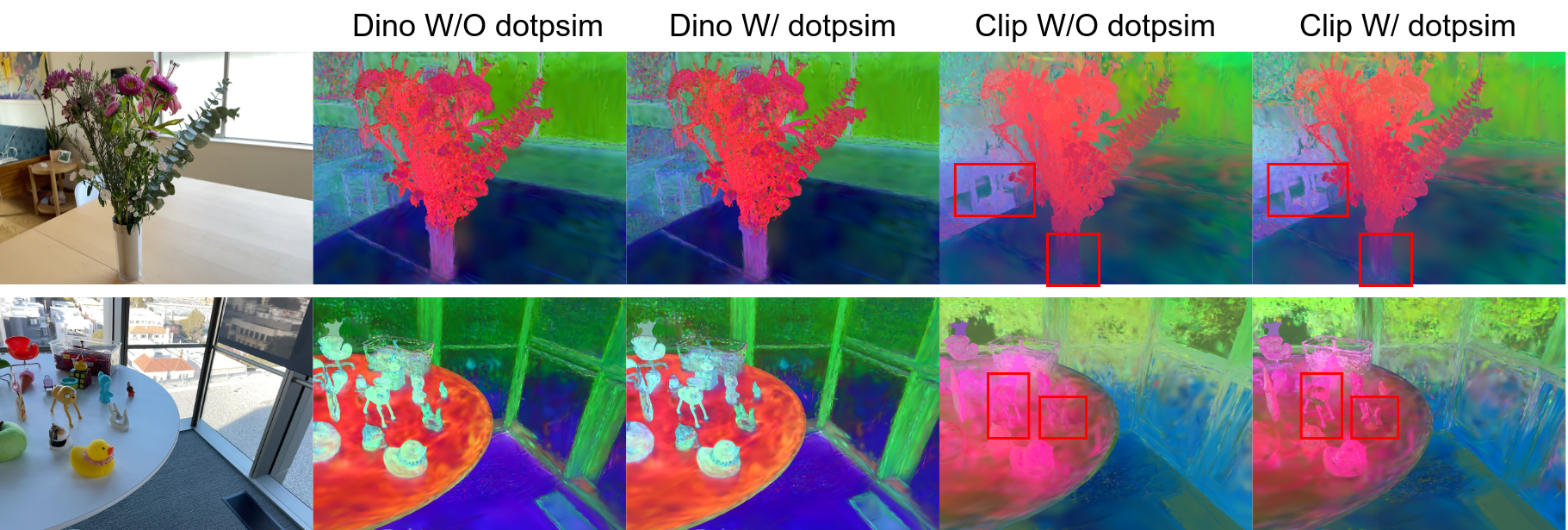}
    \vspace{-12pt}
    \caption{\textbf{Effect of dot product similarity (dotpsim) loss.} From left to right: RGB image, rendered DINO feature without dotpsim, rendered DINO feature with dotpsim, rendered CLIP without dotpsim, and rendered CLIP feature map with dotpsim. \rev{The DINO feature maps do not have significant differences with or without dotpsim.} From the CLIP feature maps, we can see that objects can be further distinguished from each other and the background. Differences are highlighted in the red boxes.}
    \label{fig:dotp_effect}
\end{figure*}

\subsection{Relevancy Score}\label{sec:relevancy}

% At query time, given a query prompt and a viewing direction, \algname{} generates a relevancy map that has a high score in semantically relevant locations (see Figure~\ref{fig:query}). To acquire this relevancy map we first render the feature map $\featureMap$ using our learned vision-language feature field. We then calculate the query's CLIP embedding $\clipOutput_{query}$. We follow \cite{kerr2023lerf} and define a set of canonical phrases with CLIP embeddings $\clipOutput_{can}$. Then we calculate a pairwise softmax between the cosine similarity of the prompt embedding with $\clipOutput_{i,j}$, the $\featureMap$ at location $(i, j)$,  and the canonical embeddings for canonical phrases.

{
At query time, when provided with a query prompt and a viewing direction, \algname{} generates a relevancy map that assigns high scores to semantically relevant locations (see Figure~\ref{fig:query}). To obtain this relevancy map, we first render the feature map $\hat{\featureMap}$ using our learned semantic feature field \rev{via GS rasterization}. Then, we calculate the CLIP embedding $\clipOutput_{query}$ corresponding to the query prompt.

\rev{To obtain the dense relevancy map}, we define a set of canonical phrases with CLIP embeddings $ \mathcal{F}_{can}$ following the methodology similar to \cite{kerr2023lerf}. Then, we compute pairwise softmax scores based on the cosine similarity between the prompt embedding and $\hat{\clipOutput}_{i,j}$, representing the $\hat{\featureMap}$ at location $(i, j)$, as well as the canonical embeddings for canonical phrases.
}
We take the minimum value of the softmax over all canonical prompts and deem it the relevancy score $r$:
\rev{
\begin{equation}
\scalemath{.9}{
    r_{i, j} = \min_{n} \frac{\exp(\hat{\clipOutput}_{i,j}^T \mathbf{f}_{query})}{\exp(\hat{\clipOutput}_{i,j}^T \mathbf{f}_{query}) + \exp(\hat{\clipOutput}_{i,j}^T \clipOutput^n_{can})},  \clipOutput^n_{can} \in \mathcal{F}_{can}}
\end{equation}}
\rev{With the above definition, the relevancy score is higher when a query embedding is closer to the rendered feature than the canonical features.} We follow \cite{kerr2023lerf} and choose the following canonical prompts: ``object", ``stuff", ``things", and ``texture". We also find that these work well for a wide range of queries removing the need for tuning these canonical terms.
\rev{In Figure~\ref{fig:raw_CLIP_relevancy}, we present representative relevancy maps generated by matching the query embedding with our rendered CLIP feature map and the target CLIP feature map from the foundation model used in our training. It's evident that the relevancy map derived from our rendered CLIP feature map overall exhibits finer granularity and higher quality.}

% We do not compute the relevancy scores on parts of the scene that did not appear in training views often. These includes the borders of the scene and positions in the image background. To remedy this during querying we discard samples that have been seen less than five times during training.

\subsection{Implementation Details}~\label{sec:implementation}
%  The hash grid used for representing language features is much larger than a typical RGB hashgrid: it has 32 layers from a resolution of 16 to 512, with a hash table size of $2^{21}$ and feature dimension of 8. The CLIP MLP used for $\nnClip$ and $\nnDino$ have ? \todo{Rocky}hidden layers with width ?? \todo{Rocky} before the final 512 dimension CLIP output. We use the OpenCLIP~\cite{cherti2022reproducible} ViT-B/16 model trained on the LAION-2B dataset, with an image pyramid varying from $s_\text{min}=??$ to $s_\text{min}=??$ in ?? steps.  The DINO MLP for $F_\text{DINO}$ has 1 hidden layer of dimension ??.

% \looseness=-1 We use the Adam optimizer for proposal networks and fields with weight decay $10^{-9}$, with an exponential learning rate scheduler from $10^{-2}$ to $10^{-3}$ over the first ?? training steps. All models are trained to 30,000 \todo{Pouya correct the numbers} steps (\todo{Pouya ask Rocky} minutes) without the pixel alignment loss and 4000. Good results can be obtained in as few as \todo{Pouya ask Rocky} as presented in the Appendix. We train on an NVIDIA \todo{which GPU?}, which takes roughly ?? \todo{Rocky} of memory total. The $\lambda$ used in weighting CLIP loss is $0.2$, chosen empirically and ablated in Sec \ref{sec:ablations}. When computing relevancy score, we multiply similarity by $10$ as a temperature parameter within the softmax.

Our approach employs a hash grid for representing language features, which is notably larger than a typical RGB hash grid. This hash grid comprises \rev{24} layers, spanning resolutions from 16 to 512, and possesses a hash table size of \rev{$2^{20}$} with an associated feature dimension of 8. The architecture of the CLIP and DINO MLP models used for $\nnClip$ and $\nnDino$ aligns with that of LERF~\cite{kerbl20233d}.
Furthermore, we leverage the OpenCLIP~\cite{cherti2022reproducible} ViT-B/16 model, which has undergone training on the LAION-2B dataset. Notably, this model operates with an image pyramid that varies in scale from $0.05$ to $0.5$ of image size, encompassing a total of seven scales for pre-computing CLIP feature pyramid. 
%\rev{We take the average pooling of the pre-computed CLIP feature pyramid to get the final hybrid CLIP feature for training our semantic embedding field.}
\rev{The pre-computed feature pyramid is subsequently processed by average pooling to generate the final hybrid CLIP feature for training our semantics embedded field.}

% \looseness=-1 We firstly train the Vanilla Gaussian Splatting scene representation~\cite{kerbl20233d} with 30K iterations, which takes about 10 minutes for a table-scale scene. There are millions of Gaussians for representing a table-scale scene. We freeze the geometric attributes and spherical harmonics of Gaussians during the proceeding semantic embedding fields training process. To maintain limited GPU memory footprint, we select around 10 percents of Gaussians with both high opacity values and 2D radius over than 2 pixels. Only the selected Gaussians are involved in the rendering when we train the semantic embeddings.  We use the RAdam optimizer with weight decay $10^{-9}$, and with an exponential learning rate scheduler from $5*10^{-3}$ to $4*10^{-3}$ over the 4.2k training steps in total. All models are firstly trained to 2500 steps with the pixel alignment loss disabled. We train and test our method on an NVIDIA RTX A500 GPU @ 24G. The $\lambda$ used in weighting CLIP loss is $0.2$, and the weight for pixel-alignment loss is $0.01$. 

% When computing relevancy score, we multiply similarity by $10$ as a temperature parameter within the softmax.
%  lerf_optimizer = torch.optim.RAdam(lerf_model.parameters(), lr=args.fmap_lr, eps=1e-15, betas=(0.9, 0.999), weight_decay=1e-9)

\looseness=-1  Initially, we train the Vanilla Gaussian Splatting scene representation~\cite{kerbl20233d} through a total \rev{number} of 30K iterations, with approximately 10 minutes \rev{total time} for a \rev{room}-scale scene. It's worth noting that representing such a scene requires the utilization of millions of Gaussians. Subsequently, we maintain the frozen states of the geometric attributes and spherical harmonics associated with these Gaussians throughout the subsequent training process for semantic embedding fields.

To mitigate GPU memory constraints, we strategically select approximately \rev{40}\% of the Gaussians based on criteria such as high opacity values and a 2D radius \rev{of projected Gaussian} exceeding 2 pixels \rev{in at least one training view}. Only these selected Gaussians are involved in the rendering process when we train the semantic embeddings.
For optimization, we employ the RAdam optimizer with a weight decay of $10^{-9}$. We incorporate an exponential learning rate scheduler, which spans from an initial value of $5 \times 10^{-3}$ and gradually decreases to $4 \times 10^{-3}$ over the course of 4.2K training steps (\rev{after the initial 30K original GS training steps}). 
In our training regimen, all models initially undergo 2.5K steps without the pixel alignment loss being enabled. These training and testing procedures are executed on an NVIDIA RTX A5000 GPU with 24GB of GPU RAM. \rev{The semantic feature field training time with a total of 4.2K steps takes about $1.4$ hours.}
\rev{During training}, we \rev{use} weighting factors to balance the CLIP loss ($\lambda = 0.2$) and the pixel-alignment loss (\rev{$\gamma$} = $0.01$). % ($\lambda = 0.2$)

\section{Experiments}
Our hybrid semantic scene representation, \algname{}, seamlessly integrates the 3D Gaussians and multi-resolution hashing encoding and supports both photo-realistic rendering and open-vocabulary object detection. In this section, we carefully evaluate the performance of open-vocabulary object detection (or localization) of our proposed method in uncontrolled real-world scenarios.  To showcase the embedding quality of our method, we also evaluate it out-of-the-box on the open-vocabulary semantic segmentation task. We compare our method to other SOTA approaches for each experiment and show significant improvement over their results.

%open-vocabulary object recognition and scene understanding. We exemplify it with two typical applications: detection of open-vocabulary objects in uncontrolled real-world scenarios and unsupervised semantic segmentation.
\subsection{Object Detection in the Wild}

\begin{table}[t]
\centering
% \small % Adjust the font size if needed
\setlength{\tabcolsep}{1pt} % Adjust the column separation
\resizebox{\linewidth}{!}{
% \begin{tabular}{lcccc}
\begin{tabular}{l *{4}{c}}  %{l{2.0cm} *{4}{c{2.0cm}}}
\toprule 
\textbf{Scene} & \textbf{FFD-LSeg}~\cite{kobayashi2022distilledfeaturefields}& \textbf{OWL-ViT}~\cite{minderer2022simple_owlvit} &  \textbf{LERF}~\cite{kerr2023lerf} & \textbf{Ours} \\
\midrule
% Scene 1 & 10.0\% & 90.0\% \\
bouquet         & 50.0\% & 66.7\% & \cellcolor{secondbest}83.3\% & \cellcolor{best}100.0 \% \\
figurines       & 8.9\%  & 38.5\% & \cellcolor{secondbest}87.2\% & \cellcolor{best}89.7\% \\
ramen           & 15.0\% & \cellcolor{best}92.5\% & 62.5\% & \cellcolor{secondbest}90.0 \% \\
teatime         & 28.1\% & 75.0\% & \cellcolor{best}96.9\% & \cellcolor{secondbest}93.8\% \\
kitchen  & 13.0\% & 42.6\% & \cellcolor{secondbest}85.2\% & \cellcolor{best}92.6 \% \\
\midrule
\textbf{Average Acc.} & 18.0\% & 54.8\% & {\cellcolor{secondbest} 83.0}\% & \cellcolor{best} 93.2\% \\
{\textbf{Inference FPS}} & {-} & {-} & {0.1214} & {103.4}  \\
\bottomrule
% | Lerf         | 0.833333    | 0.846154  | 0.80         | 0.937500   | 0.833333         | 0.85006           |
% ours           |   1.000000 |   0.897436 |   0.900000 |   0.937500 |   0.925926 |       0.932172       |
\end{tabular}
}
\caption{\textbf{Accuracy \rev{and runtime efficiency (Frames Per Second, FPS) of object detection} with open-vocabulary queries.} comparison between Feature Fields Distillation~\cite{kobayashi2022distilledfeaturefields} using LSeg~\cite{li2022language_lseg} features (FFD-Lseg), OWL-ViT~\cite{minderer2022simple_owlvit}, LERF~\cite{kerr2023lerf} and Ours \algname{}. We highlight the \colorbox{best}{best}, \colorbox{secondbest}{second-best} accuracy scores. Please find more details on scenes and text queries for LERF dataset in~\cite{kerr2023lerf}.}
\vspace{-15pt}
\label{tab:object_detection}
\end{table}

 % 0.8333   |   0.871795 | 0.625         | 0.968750   |  0.851852  | 0.8301394           |

\begin{figure*}[tbh]
    \centering
    \includegraphics[width=\textwidth]{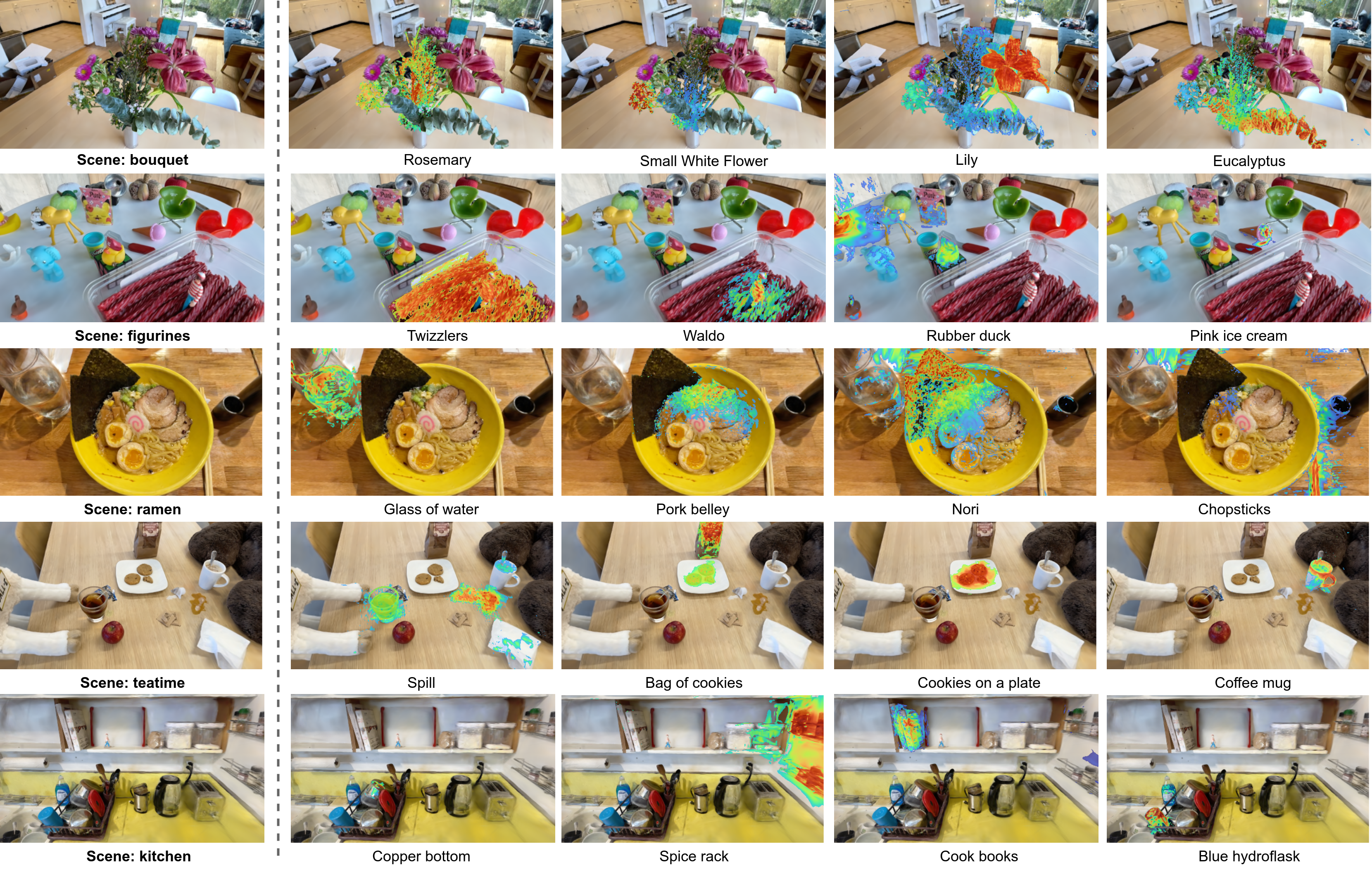}
    \caption{
    \textbf{Relevancy score for object detection.} 
    \textit{Left}: The rendered RGB image at novel view from 5 scenes on LERF dataset~\cite{kerr2023lerf}. \textit{Right}:  Visualization of relevancy scores with the given text queries shown below the figures. We \rev{overlay} them on the RGB images.}
    \vspace{-12pt}
    \label{fig:object_detection}
\end{figure*}

\begin{figure*}[tbh]
    \centering
    \includegraphics[width=\textwidth]{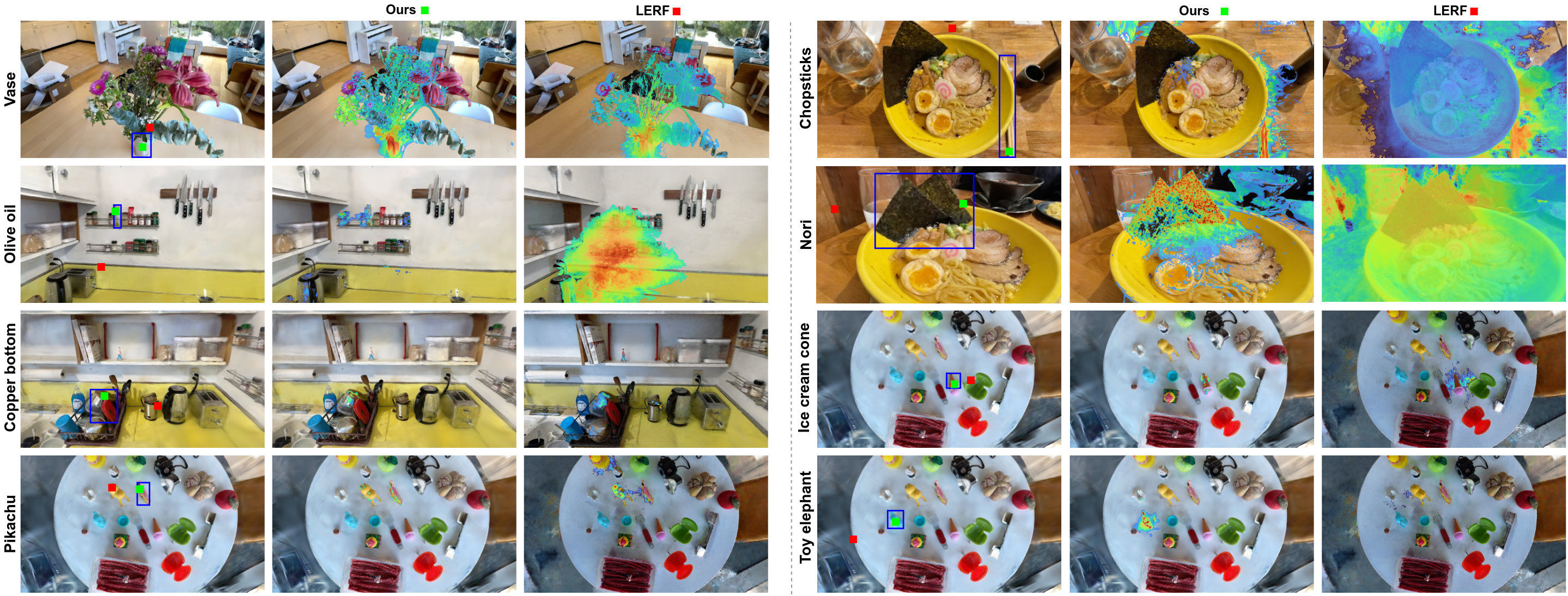}
    \caption{
    \textbf{\rev{Object detection results.}} 
    \rev{The left and right groups of figures illustrate the detection results for four open-vocabulary queries, respectively. Each group comprises three subfigures:  \textit{Left} displays the ground-truth bounding boxes (blue), our detected highest-relevancy pixel (green) and the one detected by LERF (red)~\cite{kerr2023lerf}.  \textit{Middle} showcases our relevancy score corresponding to the given text query. The text query is shown at the far left of each row. \textit{Right} showcases LERF's relevancy score corresponding to the given text query. }
    Our computed relevancy score is more focused on the target objects linked to the query.}
    \vspace{-12pt}
    \label{fig:object_detection_comparison}
\end{figure*}

\begin{table*}[thb]
\centering
\resizebox{\textwidth}{!}{
\begin{tabular}{@{}@{\,\,}c@{\,\,}|cc|cc|cc|cc|cc|cc@{}}
\toprule
{\multirow{2}{*}{Methods}}  & \multicolumn{2}{c|}{\textit{bed}} & \multicolumn{2}{c|}{\textit{sofa}} & \multicolumn{2}{c|}{\textit{lawn}} & \multicolumn{2}{c|}{\textit{room}} & \multicolumn{2}{c|}{\textit{bench}} & \multicolumn{2}{c}{\textit{table}} \\
 & \small \textbf{mIoU} & \small \textbf{mAP} & \small \textbf{mIoU} & \small \textbf{mAP} & \small \textbf{mIoU} & \small \textbf{mAP} & \small \textbf{mIoU} & \small \textbf{mAP} & \small \textbf{mIoU} & \small \textbf{mAP} & \small \textbf{mIoU} & \small \textbf{mAP} \\
\midrule
\midrule
% LSeg~\cite{li2022language_lseg} &  56.0 & 87.6 &  04.5 &  16.5 &  17.5 &  77.5 &  19.2 &  46.1 &  06.0 &  42.7 &  07.6 &  29.9 \\
OV-Seg~\cite{liang2022open_ovseg} &   79.8 &  40.4 &   66.1 &   69.6 &   81.2 &  92.1 &   71.4 &  49.1 &   88.9 &   89.2 &   80.6 &   65.3 \\
% FFD~\cite{kobayashi2022distilledfeaturefields} &  56.6 &   86.9 &  03.7 &  09.5 &  42.9 &  82.6 &  25.1 &  51.4 &  06.1 &  42.8 &  07.9 &  30.1 \\
3D-OVS ~\cite{liu2023_3dovs} &   89.5	&   96.7	&   74.0 &   91.6 &   88.2 &   97.3 &   92.8 &   98.9 &   89.3 &   96.3 &   88.8 &   96.5  \\
\midrule
\midrule
 \textbf{LERF}~\cite{kerr2023lerf} & 33.5 & 25.6& 28.1 & 45.6& 49.8 & 82.0& 26.3 & 49.1& 55.2 & 79.5&  31.1 & 33.3 \\
 \textbf{Ours} & \textbf{38.0} & \textbf{50.1} & \textbf{56.6} & \textbf{82.0} & \textbf{64.9} & \textbf{90.5} & \textbf{57.0} & \textbf{85.3} & \textbf{62.1} & \textbf{84.1} & \textbf{63.6} & \textbf{85.3} \\
 \midrule
 \midrule
 {\textbf{Ours-Refined}} & {\textbf{80.6}} & {\textbf{85.5}} & {\textbf{90.8}} & {\textbf{97.4}} & {\textbf{92.6}} & {\textbf{98.5}} & {\textbf{87.9}} & {\textbf{97.8}} & {\textbf{84.5}} & {\textbf{94.8}} & {\textbf{89.4}} & {\textbf{97.2}} \\
\bottomrule
\end{tabular}
}
\caption{
\textbf{Segmentation Evaluation.} We report the mIoU($\uparrow$) scores and the mAP($\uparrow$) scores of the following methods in 6 scenes of 3D-OVS dataset~\cite{liu2023_3dovs}. Note that 3D-OVS \rev{is a weakly supervised method}, which knows the segmentation annotations in training and specially designed for segmentation task. Our method and LERF are 3D method training without any segmentation annotations, relying only on the relevancy between class query and the rendered CLIP features. \rev{OV-Seg~\cite{liang2022open_ovseg} is a supervised method for segmentation task}. Our method and LERF are \rev{unsupervised methods}, under apple-to-apple comparison. \rev{We can further post-process and refine our 2D segmentation results by SAM~\cite{kirillov2023segment} and get the results shown as `Ours-Refined'.}
}
\label{tab:segmentation}
\end{table*}

By distilling the language embeddings extracted from off-the-shelf vision-language model, CLIP,  our \algname{} is applicable for associating a wide range of textual prompts with the relevant vision clues. We test the open-vocabulary object understanding capability of our method by object detection experiments. 

\textbf{Dataset:} We use the same dataset as used in the LERF~\cite{kerr2023lerf} for object detection evaluation, for the purpose of fair comparison. It consists of five labelled scenes with 2D bounding boxes of objects associated with text prompts.  There are objects including both common and long-tail ones with different sizes, and the queries for objects are quite diverse, like `vase', `eucalyptus', `big white crinkly flower', `pikachu', `twizzlers',  `spoon handle', `power outlet', `waldo', `stuffed bear', `cookies on a plate', etc. The location of queried images are labelled by bounding boxes in the test images, which are rendered at novel views from trained NeRF models of individual scenes. 
The scenes in LERF dataset are collected by an iPhone, and each scene comprise $\sim200$ images.
The provided poses of images from Ploycam app are with significant noises in some scenes. Thus we regenerate the poses of images by running COLMAP~\cite{schonberger2016structure_colmap}, which also \rev{yields sparse 3D visual points serving as input} to initialize 3D Gaussians in our method. The poses of the officially-provided test images are also properly transferred to our COLMAP trajectory by Sim(3) alignment between officially-provided image poses and our COLMAP poses. 

\textbf{Evaluation Protocol:} Following LERF~\cite{kerbl20233d}, the evaluation metric for object detection is the accuracy rate. We redeem the query is a success if the highest relevancy pixel locates inside the target box. The relevancy score at each pixel is obtained by matching the rendered CLIP feature map with the language embedding of given text query as described in Sec.~\ref{sec:relevancy}.

\textbf{Baselines:} We compare against FFD-LSeg that embeds pixel-aligned LSeg feature~\cite{li2022language_lseg} into NeRF (NeuralStudio `neurfacto' implementation by feature fields distillation method~\cite{kobayashi2022distilledfeaturefields},  OWL-ViT~\cite{minderer2022simple_owlvit} that is a 2D method based on Vision Transformer encoder and fine-tuned for object detection, LERF~\cite{kerr2023lerf} that embeds CLIP and DINO features into NeRF. The 3D methods, FFD-LSeg and LERF, share the same evaluation protocol as our \algname{}. For the 2D method, OWL-ViT, we regard it as a success if the center of the predicted bounding box locates in the target box.

\begin{table}[t]
\centering
% \small % Adjust the font size if needed
\setlength{\tabcolsep}{1pt} % Adjust the column separation
\resizebox{\columnwidth}{!}{
\begin{tabular}{p{2.8cm} *{5}{c}|c} % {l{2.0cm}*{4}{c}|c}
\toprule 
\textbf{Methods} & bouquet & figurines & ramen & teatime & kitchen & \textbf{Average} \\
\midrule
\textbf{Ours} & 100.0 & 89.7 & 90.0 & 93.8 & 92.6 & \textbf{93.2}\\
\midrule
\textbf{W/O dotpsim} &   100.0 & 91.0 & 85.0 & 90.6 & 85.2& 90.4\\
\textbf{W/O hybrid CLIP} & 54.2 & 32.1 & 52.5 & 6.3 & 9.3 &  30.8\\
{\textbf{W/ LERF CLIP}} & {91.7} & {70.5} & {72.5} & {72.5} & {87.0} & {78.8} \\
{\textbf{W/O MHE}} & {91.7} & {71.8} & {90.0} & {90.6} & {77.8} & {84.4} \\
\bottomrule
% | Lerf         | 0.833333    | 0.846154  | 0.80         | 0.937500   | 0.833333         | 0.85006           |
% ours           |   1.000000 |   0.897436 |   0.900000 |   0.937500 |   0.925926 |       0.932172       |
\end{tabular}
}
\caption{\textbf{Ablation study.} Object detection comparison between our full method, ours without dot product similarity (dotpsim) loss, and ours without hybrid CLIP features by averaging at multiple scales for supervision, using single scale CLIP feature at the finest-resolution instead,\rev{ as well as ours with LERF CLIP at multiple individual scales~\cite{kerr2023lerf}}.}
\vspace{-15pt}
\label{tab:ablation}
\end{table}

\textbf{Evaluation Results:} The quantitative evaluation results on all sequences of LERF dataset are presented in Table~\ref{tab:object_detection}, and representative relevancy score maps of the proposed method are shown in Figure~\ref{fig:object_detection}. The detailed results demonstrate significant advantages of \algname{}'s integration of language embeddings in detecting objects associated with long-tail prompts.
While LSeg~\cite{li2022language_lseg}, trained on a small dataset to learn pixel-aligned CLIP features, exhibits diminished open-vocabulary language understanding capabilities, the approach of FFD-LSeg, which distills LSeg features into radiance fields, struggles with comprehending long-tail queries and consequently exhibits poorer performance.
In terms of open-vocabulary 2D detection, Owl-ViT, which utilizes full-HD NeRF views and selects bounding boxes based on the highest confidence scores for text queries, outperforms FFD-Lseg. However, when faced with long-tail queries, Owl-ViT's performance falls short in comparison to the robust and versatile \algname{}.

We also conducted a comparison with the closest method, LERF, which distills DINO and CLIP features into neural radiance fields represented solely by MHEs. % \rev{and MLPs.}
As depicted in Table~\ref{tab:object_detection}, our \algname{} outperforms LERF significantly, achieving an accuracy improvement of $\improvement$ percentage points. Note that our tested LERF results, obtained using the officially released code, slightly surpasses those reported in the original paper~\cite{kerr2023lerf}.

In Figure~\ref{fig:object_detection_comparison}, we present side-by-side comparisons with LERF~\cite{kerr2023lerf}. The object detection results are visualized, highlighting the superior quality of the relevance map produced by our \algname{}. It notably focuses more on the queried target objects, as opposed to LERF. This outcome stems from our hybrid representation, which combines 3D Gaussians and MHEs for semantic scene representation. The 3D Gaussians represent both the geometry and appearance of the scene, naturally dividing 3D structures of objects and the scene into distinct Gaussian volumes. This partitioning feature aids in distinguishing objects from each other and from the background. In \algname{}, we assign an identical MHE embedding to a Gaussian volume, further promoting semantic consistency in local proximity. This, in turn, contributes to the focusing of relevance on the target object.
Taking the query `Pikachu' in Figure~\ref{fig:object_detection_comparison} as an example, where `Pikachu' is depicted on the side of a paper bag. Even when observing from a challenging viewpoint with almost no visibility of `Pikachu', \algname{} successfully maintains high relevance at the target location, due to its 3D consistency and fine-grained scene understanding. In contrast, LERF fails to detect `Pikachu' and mistakenly identifies a visually similar object.

% Our \algname{} relying on 3D Gaussain Splatting render~\cite{kerbl20233d} is super efficient at rendering RGB images. We implemented our render for the CLIP and DINO feature maps based on Cuda implementation of Gaussain Splatting render. Although render deep features with high dimensions significantly increase the computation time, our \algname{} is still super fast, rendering the $480 \times 270$ CLIP feature map, DINO feature map, and RGB image jointly at  $103.4$ FPS during inference.  In contract, LERF runs at $0.1214$ FPS during inference, since it needs to select the best scales when rendering CLIP features despite its render is slower. LERF performs the brute force search of the scale across a range of scales 0 to 2 meters with 30 increments. There fore, we are \textbf{851.73} times faster than LERF for rendering clip features in order to achieve open-vocabulary query.

\textbf{Inference Runtime:} Our \algname{}, relying on 3D Gaussian Splatting rendering~\cite{kerbl20233d}, excels in efficiently rendering RGB images. We've implemented our rendering method for CLIP and DINO feature maps based on a CUDA implementation of Gaussian Splatting rendering. Even when rendering deep features with high dimensions, which can significantly increase computation time, our \algname{} remains remarkably fast. It can render the $480 \times 270$ CLIP feature map, DINO feature map, and RGB image jointly at an impressively high rate of $103.4$ FPS during inference, \rev{even with our unoptimized implementation}. In contrast, LERF operates at a significantly slower pace, achieving a mere $0.1214$ FPS during inference (\rev{see Table~\ref{tab:object_detection}}). This slowness stems from LERF's need to perform a brute-force search for the best scales when rendering CLIP features, spanning a range from 0 to 2 meters with 30 increments. Consequently, we are \textbf{851.73 times faster} than LERF in rendering CLIP features, enabling efficient real-time open-vocabulary queries after our scene representation is trained.

% we generate relevancy maps across a range of scales 0 to 2 meters with 30 increments, and select the scale that yields the highest relevancy score.

\begin{figure*}[tbh]
    \centering
    \includegraphics[width=\textwidth]{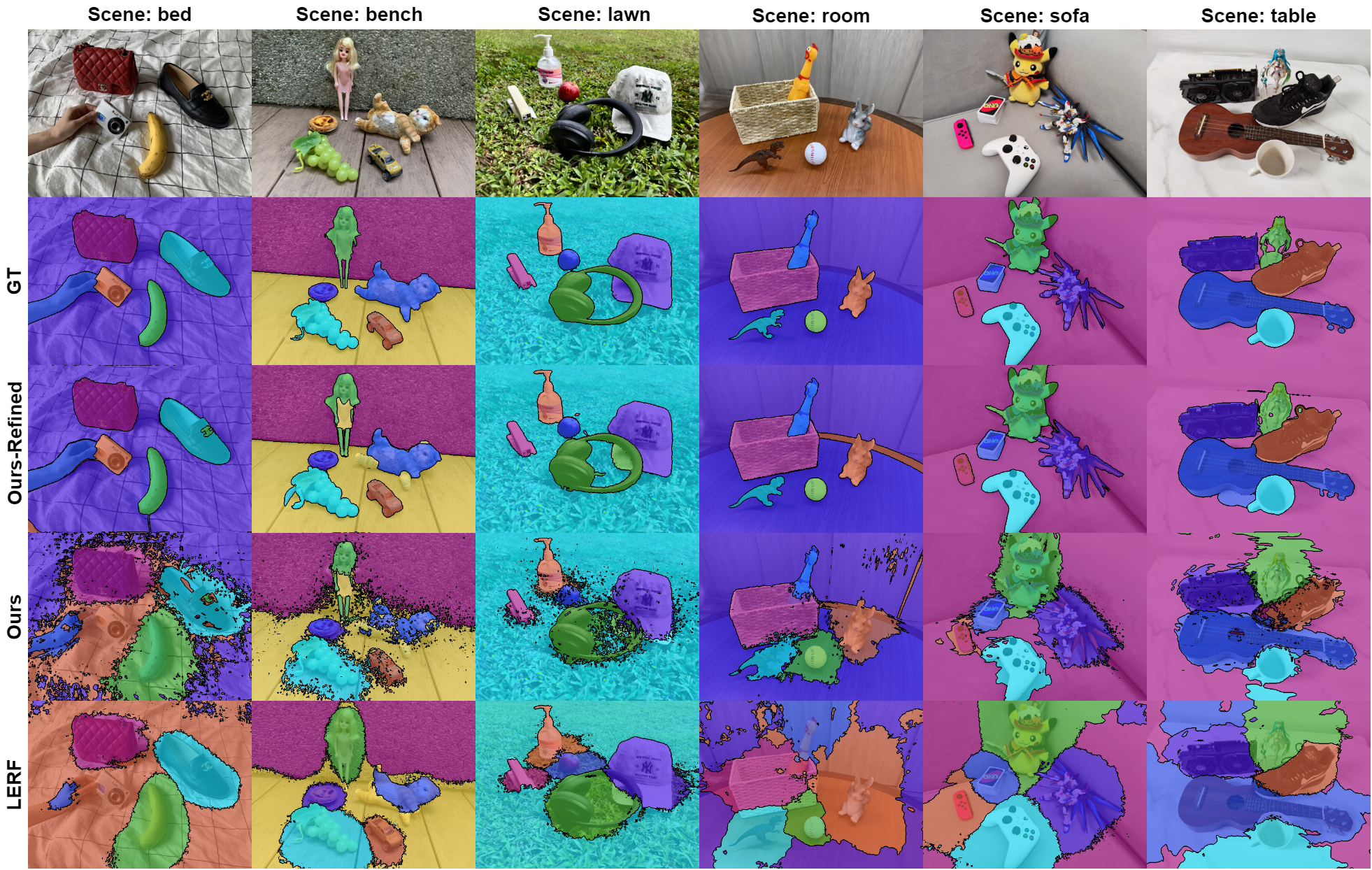} %{figures/le3gs_segmentation.png}
    \caption{
    \textbf{Semantic segmentation results.} 
   In the rows from top to bottom, we display RGB images, ground-truth (GT) segmentation masks, \rev{our refined segmentation results,} our segmentation results, and the segmentation results obtained by LERF~\cite{kerr2023lerf} scene representation. % It's important to highlight that both our method and LERF were not originally designed for the segmentation task. Instead, our objective is to assess the quality of the relevance map computed from the rendered CLIP features.
    It's essential to note that neither our method nor LERF was initially intended for the segmentation task. Our primary aim is to evaluate the pixel accuracy of the relevance map computed from the rendered CLIP features. \rev{We can further post-process and refine our 2D segmentation results by SAM~\cite{kirillov2023segment} and get the results shown as `Ours-Refined'.}
    }
    \vspace{-12pt}
    \label{fig:segmentation}
\end{figure*}

\subsection{Unsupervised Segmentation}

In \rev{following} experiments we use \algname{} to segment queries and evaluate their segmentation masks. \textbf{Note} that our method is not delicately designed for the segmentation task. We lack a dedicated segmentation header for predicting segmentation masks, nor do we explicitly partition the scene at the object level. We have examined the open-vocabulary language understanding capability of \algname{} in the object detection experiments discussed in the above section. 
%
% The purpose of conducting this segmentation task is mainly for examining the pixel-accuracy of rendered CLIP features from the trained scene representation, since the segmentation relies on matching the rendered CLIP features to the embeddings of the given semantic labels. 
%
Our primary objective for doing this segmentation evaluation is to assess the pixel-level accuracy of the rendered CLIP features obtained from the trained scene representation. Segmentation relies on matching these rendered CLIP features to the embeddings of the provided semantic labels.

% We do not have any knowledge of the known semantic labels during our training process.

% We use a dataset that encompasses 10 unique scenes. Within each scene resides an assortment of objects categorized as ``long-tail," implying their infrequent occurrence. These objects are positioned in diverse poses against a variety of backdrops to simulate real-world scenarios.

\textbf{Dataset:}  We conducted our segmentation evaluation using the 3D-OVS dataset~\cite{liu2023_3dovs}, which consists of six scenes with labeled ground-truth semantic segmentation masks for test image views. These scenes are characterized by their cleanliness, with clear backgrounds and well-defined foreground objects. Each scene comprises approximately 30 images with predefined poses and sparse points computed using COLMAP~\cite{schonberger2016structure_colmap}. The dataset includes a variety of objects, including many long-tail objects like `Gundam,' `Pikachu,' `stapler', and more. For further details about the scenes and semantic labels, please refer to~\cite{liu2023_3dovs}.

\textbf{Evaluation Protocol:} In terms of our evaluation protocol, we rely on the annotated ground-truth masks for the test views. These masks serve as a reliable benchmark for both qualitative and quantitative assessments of segmentation performance.
We calculate the mean Intersection over Union (mIOU) scores and mean Average Precision (AP) metrics by comparing the segmentation results with these ground-truth masks.

\textbf{Baselines:} We conduct a direct comparison of our method with LERF~\cite{kerr2023lerf}. To perform semantic segmentation, we initially obtain relevancy scores by computing the cosine similarity between the rendered CLIP feature and the embeddings of all class labels \rev{(this is different from the relevancy score calculation with auxiliary canonical phrases involved in Sec.~\ref{sec:relevancy}.)}. These relevancy scores serve as segmentation logits, and we subsequently apply the softmax function to convert them into probabilities. Each pixel is then assigned a semantic class label corresponding to the maximum probability. 
Note that LERF~\cite{kerr2023lerf} requires a scale factor when rendering CLIP features, and we report the best segmentation results that can be achieved by LERF by selecting the best scales for each ray.
\rev{It's also important to note that both LERF and our method encounter challenges in discerning the semantic labels of backgrounds when presented with visibility-limited close views and lack of context. Therefore, we have replaced the original background labels, including `white sheet', `wood wall', `grey sofa', and `lime wall', with a more general label `background' when testing LERF and our method.}
% Note that our method and LERF struggle to recognize the semantic meanings of the backgrounds with only seeing a  very limited view of them, thus we replace the original background labels, `white sheet', `wood wall',  `grey sofa' and `lime wall' with `background'.

Additionally, for comprehensive reference, we present results obtained using the dedicated 3D-OVS method~\cite{liu2023_3dovs} for the segmentation task. However, it is worth emphasizing that comparing object detection methods like ours and LERF~\cite{kerr2023lerf} to 3D-OVS is not entirely equitable, as acknowledged in the paper of 3D-OVS~\cite{liu2023_3dovs}.
3D-OVS~\cite{liu2023_3dovs} has prior access to segmentation class labels and distill class-related information into the radiance field during training.
In contrast, neither LERF nor our methods have access to class labels during scene representation training. Consequently, the trained 3D-OVS scene representation can only be effectively employed for querying the classes known before training, and does not support arbitrary semantic queries beyond the trained classes.
Furthermore, we compare to a 2D ceiling approach \cite{liang2022open_ovseg}, OV-Seg, which is directly trained for open-vocabulary semantic segmentation by fine-tuning CLIP on masked image regions and text descriptions. \rev{OV-Seg is supervised with mask-category pairs, while ours and LERF are completely unsupervised.}

\textbf{Evaluation Results} The segmentation experiment results are presented in Table~\ref{tab:segmentation} and Figure~\ref{fig:segmentation}. Notably, our approach outperforms LERF~\cite{kerr2023lerf} by a significant margin across all cases. This superior performance can be attributed to the higher quality of our rendered CLIP feature compared to the one produced by LERF. Our method exhibits more concentrated high relevancy around the queried objects, showcasing the advantage of our semantic scene representation, which maintains high semantic consistency in local proximity. \rev{We refine our 2D segmentation results using SAM~\cite{kirillov2023segment} by assigning class labels to SAM segments through a majority voting based on pixel labels obtained from our method. The refined result is displayed in the `Ours-Refined' row of Table~\ref{tab:segmentation} and Figure~\ref{fig:segmentation}.}

% It is important to emphasize that 3D-OVS~\cite{liu2023weakly} requires knowing the query classes before training the language field, making it challenging to provide a direct comparison with our approach. However, we believe that similar training strategies can be applied to our \algname{} backbone, enabling a more direct comparison in future experiments focused on priorily-known class segmentation.

\subsection{Ablations}

We conducted an ablation study on the object detection task, as it serves as a key indicator of our method's open-vocabulary semantic understanding capabilities. The results are presented in Table~\ref{tab:ablation}.

\subsubsection{\rev{Hybrid CLIP feature}} 
 % When supervising the training of the scene representation with CLIP features, instead of using hybrid CLIP features obtained by averaging multiple-scale CLIP features extracted from patches at different resolution,  we use a single scale of the CLIP features extracted at the finest-resolution in this ablation study. As shown in Table~\ref{tab:ablation}, our hybrid CLIP feature has a large positive impact on the results. The scene understanding capability of our method is disasterly destroyed when using  a single-scale CLIP features.

In this ablation study, we investigated using a single scale of CLIP features \rev{at the finest scale level}, rather than our hybrid CLIP features, which are obtained by averaging multiple-scale CLIP features extracted from patches at different resolutions. As demonstrated in Table~\ref{tab:ablation}, the hybrid CLIP feature for supervision is greatly important. The scene understanding capability is severely compromised when employing only a single-scale CLIP feature for supervision \rev{(denoted as \textit{`W/O Hybrid CLIP'})}. 
\rev{The inferior performance observed with the use of a single-scale CLIP feature stems from its potential inadequacy in capturing sufficient contextual information within the image. By contrast, our hybrid CLIP feature encompasses a significantly larger receptive field, enhancing its contextual awareness.}

\rev{
To conduct a comprehensive analysis, we also compare our method of using a single hybrid CLIP supervision (\textit{`Ours'}) to the approach of utilizing CLIP supervision at multiple individual scales, as proposed by LERF~\cite{kerr2023lerf}, which we denote as \textit{`W/ LERF CLIP'}.
During its training, randomly sampled scale factors are concatenated with the intermediate MHE feature vector $\mathbf{q}$ (refer to Sec.~\ref{sec:ffa}) to decode CLIP features at each scale. At inference time, the optimal scale factors at each pixel location are determined by an exhaustive search among 30 uniformly distributed scales. The best scale factor is selected based on its relevance score to the query. Consequently, this approach is at least 30 times slower than our method with a single hybrid CLIP supervision during inference. Despite its inefficiency, the accuracy achieved by \textit{` W/ LERF CLIP'} ($78.8\%$) significantly lags behind that of \textit{`Ours'} ($93.2\%$), as shown in Table~\ref{tab:ablation}, underscoring the efficacy and superior efficiency of our proposed method. It is worth noting that while it is plausible that `W/ LERF CLIP' could achieve better performance with much more training iterations, this aspect falls beyond the scope of our current investigation.}

%\rev{For a thorough study, we also compare to using CLIP supervision at multiple individual scales (denoted as \textit{`Ours W/ LERF CLIP'}), following LERF~\cite{kerr2023lerf}. During its training time, randomly sampled scale factors are concatenated with the intermediate MHE feature vector $\mathbf{q}$ (see Sec.~\ref{sec:ffa}) for decoding CLIP features at the scale. During the reference time, the best scale factors at each pixel location are searched among 30 uniformly distributed scales in a brute-force way. The best scale factor is the one with the highest relevance score to the query. Therefore, it is at least 30 times slower than our method with a single hybrid CLIP supervision (denoted as \textit{`Ours'}) at inference. Despite its inefficiency, the accuracy of \textit{`Ours W/ LERF CLIP'} $78.8$ is much inferior to Ours $93.2$ in Table~\ref{tab:ablation}, which demonstrates the effectiveness and high efficiency of utilizing hybrid CLIP supervision in our proposed method. We suspect that \textit{`Ours W/ LERF CLIP'} with much more training iterations can have a better performance, however, this is out of our investigation scope.}

\subsubsection{Pixel-alignment loss}
% To show the effectiveness of our proposed pixel alignment loss, we train our semantic scene representation without this loss for ablation study. Its impact on the accuracy of object detection task is shown in Table \ref{tab:ablation}. We further show  the qualitative results in Figure~\ref{fig:raw_clip_relevancy}, which underscore that the CLip features from scene representation trained with pixel-alignment loss can better distinguish different objects and the object from the back ground. 

To assess the effectiveness of our proposed pixel alignment loss, we conducted an ablation study by training our semantic scene representation without this loss. The impact of omitting the pixel alignment loss on the accuracy of the object detection task is shown in Table~\ref{tab:ablation}. Furthermore, we provide qualitative results in Figure~\ref{fig:dotp_effect}, which indicates that CLIP features from a scene representation trained with pixel-alignment loss are better at distinguishing between different objects and separating objects from the background.

\subsubsection{\rev{Scene Representation}}

\rev{
To evaluate the effectiveness of our hybrid scene presentation, which integrates 3D Gaussians for geometry and appearance representation alongside MHE for efficient semantic embedding, we compare it with a vanilla scene representation method that solely employs Gaussians without MHE by attaching semantic embeddings to each Gaussian (\textit{`W/O MHE'}). Notably, for fair comparisons, we consider only the selected Gaussians with sufficient opacity and 2D radius in the vanilla method (as detailed in Sec.~\ref{sec:implementation}) identical to our method (\textit{`Ours'}). These attached semantic embeddings share the same dimensionality as the intermediate MHE feature vector $\mathbf{q}$ in our method (see Sec.~\ref{sec:ffa}). They are also decoded by two MLPs, $\nnClip$ and $\nnDino$, to obtain CLIP and DINO features while rendering the feature maps.}

\rev{
As indicated in Table~\ref{tab:ablation}, \textit{`W/O MHE'} achieves an object detection accuracy of 84.4\%, which is inferior to \textit{`Ours'} with its hybrid scene representation. The superior performance of \textit{`Ours'} can be attributed to its ability to render CLIP feature maps at a higher quality over \textit{`W/O MHE'}. Additionally, in terms of scene representation complexity, the MHE component of our method maintains a constant number of parameters across the five scenes of the LERF dataset, whereas the parameter count of the semantic embeddings in \textit{`W/O MHE'} varies with the number of Gaussians and increases by 84.1\%, 113.5\%, 2.14\%, 174.2\%, and 136.9\% compared to \textit{`Ours'}. This substantial memory demand of \textit{`W/O MHE'} can pose even greater challenges in large-scale scenarios.}
\section{Discussion and Limitations}
When comparing \algname{} to LERF\cite{kerr2023lerf}, both methods distill Clip and Dino features from foundation models into 3D scene representations. However, their rendering algorithms and scene representations differ significantly. These distinctions lead to rapid and high-quality language feature acquisition using common hyperparameters, such as the feature field architecture. An additional key advantage of \algname{} is that it employs the same feature embedding for each Gaussian, regardless of the viewing direction. This feature enables direct 3D localization of vision-language queries. It's important to note that \algname{} not only facilitates the localization of language queries in 3D but also allows for finding a given image of the scene using the 3D Gaussian embeddings. LERF, on the other hand, does not offer such 3D localization capabilities out of the box.

In terms of limitations, \algname{} currently relies heavily on the presence of high-quality and calibrated input images, a limitation shared with NeRF-based approaches. 
% Additionally, \algname{} does not handle Gaussians that could be used to render multiple objects from different viewpoints, although this scenario is relatively infrequent in less cluttered scenes.
% \rev
Additionally, the performance of \algname{} is entirely contingent on the quality of the base foundation models used for training the feature fields. It is conceivable that a model better suited for localizing language within images could yield improved feature field quality. 
\rev{Furthermore, to enhance performance in semantic segmentation tasks, it is advisable to embed a specialized segmentation foundation model, such as SAM~\cite{kirillov2023segment, cen2023segment}, into our scene representation. Unlike using SAM solely for post-refinement of our 2D segmentation results, directly distilling SAM results into the 3D space offers the advantage of enforcing multi-view consistency, potentially leading to improved performance.}

%  Alternatively, a straightforward approach for semantic segmentation is to initially segment the images using the foundation model and then assign semantic meanings to the segments based on our rendered CLIP features.

\section{Conclusions}
\algfull{} (\algname{}) contributes to scene understanding by seamlessly merging vision-language embeddings and 3D representation. This novel 3D scene representation achieves multi-view semantic consistency through self-supervised distillation and pixel alignment of CLIP features. The resulting feature embedded 3D Gaussians achieve state-of-the-art performance in comparison to previous methods. By bridging vision, language, and 3D, \algname{} paves the way for unprecedented object comprehension in real-world environments, opening exciting possibilities for augmented reality, robotics, and beyond.

\section{Acknowledgement}
We are very grateful to Juan J. Gómez Rodríguez and Francis Engelmann for their advice and insightful discussions about this work.

% \input{sections/5_appendix}

% references
% {\small
% \bibliographystyle{unsrtnat}
% }
\newpage



\bibliography{main}
\end{document}